\def\year{2018}\relax
\documentclass[letterpaper]{article}
\usepackage{aaai18}
\usepackage{times}
\usepackage{helvet}
\usepackage{courier}
\usepackage{url}
\usepackage{graphicx}
\usepackage{subfigure} 
\usepackage{makecell}
\usepackage{microtype}
\usepackage{amsmath}
\usepackage{amsthm}
\usepackage{amssymb}
\usepackage{amsfonts}
\usepackage{bm}
\begin{document}

\title{Multi-Adversarial Domain Adaptation\thanks{Corresponding author: Mingsheng Long}}

\author{
   Zhongyi Pei\thanks{Equal contribution}, Zhangjie Cao\footnotemark[2], Mingsheng Long, and Jianmin Wang\\
   KLiss, MOE; NEL-BDS; TNList; School of Software, Tsinghua University, China\\
   {\tt\small \{peizhyi,caozhangjie14\}@gmail.com \ \{mingsheng,jimwang\}@tsinghua.edu.cn}
}

\maketitle

\begin{abstract}
Recent advances in deep domain adaptation reveal that adversarial learning can be embedded into deep networks to learn transferable features that reduce distribution discrepancy between the source and target domains. Existing domain adversarial adaptation methods based on single domain discriminator only align the source and target data distributions without exploiting the complex multimode structures. In this paper, we present a multi-adversarial domain adaptation (MADA) approach, which captures multimode structures to enable fine-grained alignment of different data distributions based on multiple domain discriminators. The adaptation can be achieved by stochastic gradient descent with the gradients computed by back-propagation in linear-time. Empirical evidence demonstrates that the proposed model outperforms state of the art methods on standard domain adaptation datasets.
\end{abstract}

\section{Introduction}
Deep networks, when trained on large-scale labeled datasets, can learn transferable representations which are generically useful across diverse tasks and application domains \cite{cite:ICML14DeCAF,cite:NIPS14CNN}. However, due to a phenomenon known as dataset bias or domain shift \cite{cite:CVPR11DB}, predictive models trained with these deep representations on one large dataset do not generalize well to novel datasets and tasks. The typical solution is to further fine-tune these networks on task-specific datasets, however, it is often prohibitively expensive to collect enough labeled data to properly fine-tune the high-capacity deep networks. Hence, there is strong motivation to establishing effective algorithms to reduce the labeling consumption by leveraging readily-available labeled data from a different but related source domain. This promising transfer learning paradigm, however, suffers from the shift in data distributions across different domains, which poses a major obstacle in adapting classification models to target tasks \cite{cite:TKDE10TLSurvey}. 

Existing transfer learning methods assume shared label space and different feature distributions across the source and target domains. These methods bridge different domains by learning domain-invariant feature representations without using target labels, and the classifier learned from source domain can be directly applied to target domain. Recent studies have revealed that deep neural networks can learn more transferable features for domain adaptation \cite{cite:ICML14DeCAF,cite:NIPS14CNN}, by disentangling explanatory factors of variations behind domains. The latest advances have been achieved by embedding domain adaptation modules in the pipeline of deep feature learning to extract domain-invariant representations \cite{cite:Arxiv14DDC,cite:ICML15DAN,cite:ICML15RevGrad,cite:ICCV15SDT,cite:NIPS16RTN,cite:NIPS16SDSN,cite:ICML17JAN}.

\begin{figure*}[tbp]
  \centering
  \includegraphics[width=0.8\textwidth]{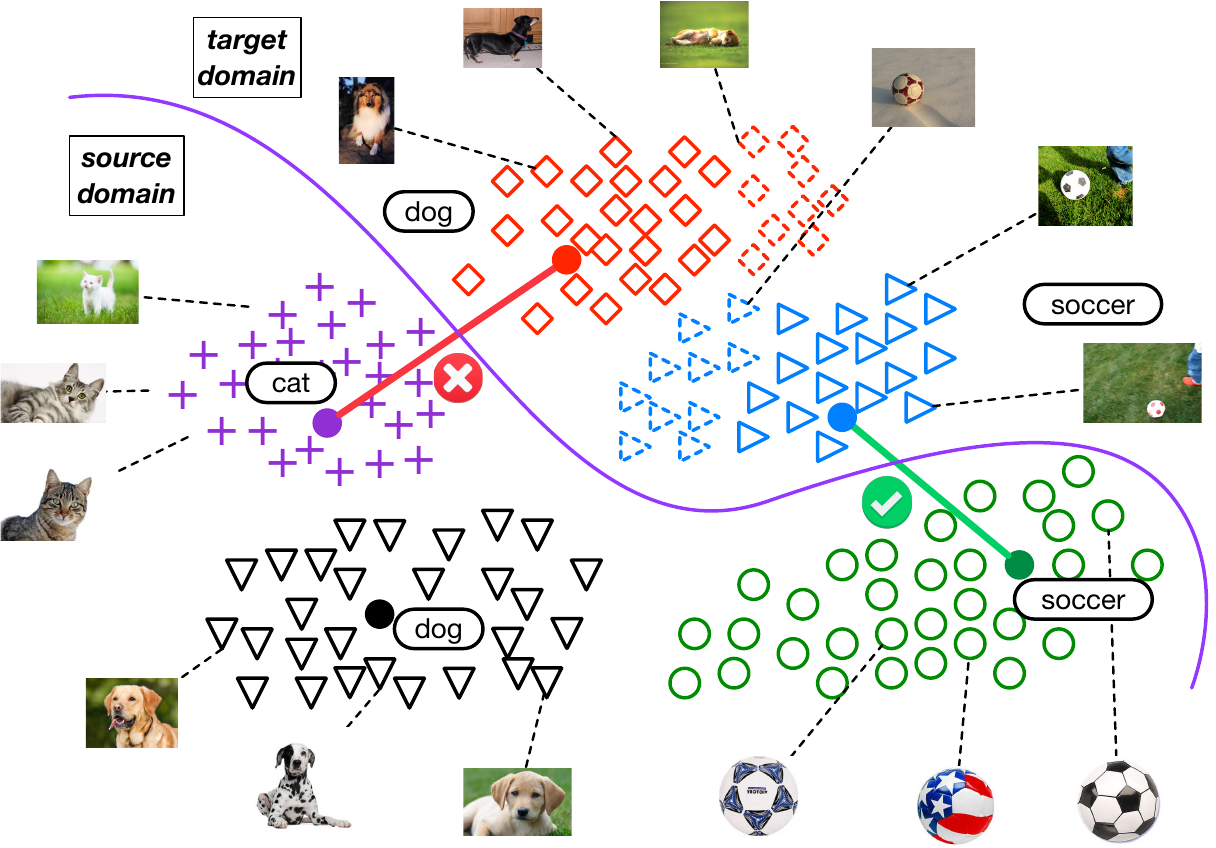}
  \caption{The difficulty of domain adaptation: discriminative structures may be mixed up or falsely aligned across domains. As an intuitive example, in this figure, the source class \textit{cat} is falsely aligned with target class \textit{dog}, making final classification wrong.}
   \label{fig:MADAproblem}
\end{figure*}

Recently, adversarial learning has been successfully embedded into deep networks to learn transferable features to reduce distribution discrepancy between the source and target domains. Domain adversarial adaptation methods \cite{cite:ICML15RevGrad,cite:ICCV15SDT} are among the top-performing deep architectures. These methods mainly align the whole source and target distributions, without considering the complex multimode structures underlying the data distributions. As a result, not only all data from the source and target domains will be confused, but also the discriminative structures could be mixed up, leading to false alignment of the corresponding discriminative structures of different distributions, with intuitive example shown in Figure~\ref{fig:MADAproblem}. Hence, matching the whole source and target domains as previous methods without exploiting the discriminative structures may not work well for diverse domain adaptation scenarios. 

There are two technical challenges to enabling domain adaptation: {(1)} enhancing \emph{positive} transfer by maximally matching the multimode structures underlying data distributions across domains, and {(2)} alleviating \emph{negative} transfer by preventing false alignment of modes in different distributions across domains. Motivated by these challenges, we present a multi-adversarial domain adaptation (MADA) approach, which captures multimode structures to enable fine-grained alignment of different data distributions based on multiple domain discriminators. A key improvement over previous methods is the capability to simultaneously promote positive transfer of relevant data and alleviate negative transfer of irrelevant data. The adaptation can be achieved by stochastic gradient descent with the gradients computed by back-propagation in linear-time. Empirical evidence demonstrates that the proposed MADA approach outperforms state of the art methods on standard domain adaptation benchmarks.

\section{Related Work}
Transfer learning \cite{cite:TKDE10TLSurvey} bridges different domains or tasks to mitigate the burden of manual labeling for machine learning~\cite{cite:TNN11TCA,cite:TPAMI12DTMKL,cite:ICML13TCS,cite:NIPS14FTL}, computer vision \cite{cite:ECCV10Office,cite:CVPR12GFK,cite:NIPS14LSDA} and natural language processing \cite{cite:JMLR11MTLNLP}. The main technical difficulty of transfer learning is to formally reduce the distribution discrepancy across different domains. Deep networks can learn abstract representations that disentangle different explanatory factors of variations behind data \cite{cite:TPAMI13DLSurvey} and manifest invariant factors underlying different populations that transfer well from original tasks to similar novel tasks \cite{cite:NIPS14CNN}. Thus deep networks have been explored for transfer learning \cite{cite:ICML11DADL,cite:CVPR13MidLevel,cite:NIPS14LSDA}, multimodal and multi-task learning \cite{cite:JMLR11MTLNLP,cite:ICML11MDL}, where significant performance gains have been witnessed against prior shallow transfer learning methods. 

However, recent advances show that deep networks can learn abstract feature representations that can only reduce, but not remove, the cross-domain discrepancy \cite{cite:ICML11DADL,cite:Arxiv14DDC}, resulting in unbounded risk for target tasks \cite{cite:COLT09DAT,cite:ML10DAT}. Some recent work bridges deep learning and domain adaptation \cite{cite:Arxiv14DDC,cite:ICML15DAN,cite:ICML15RevGrad,cite:ICCV15SDT,cite:NIPS16RTN,cite:NIPS16SDSN,cite:ICML17JAN}, which extends deep convolutional networks (CNNs) to domain adaptation by adding adaptation layers through which the mean embeddings of distributions are matched \cite{cite:Arxiv14DDC,cite:ICML15DAN,cite:NIPS16RTN}, or by adding a subnetwork as domain discriminator while the deep features are learned to confuse the discriminator in a domain-adversarial training paradigm \cite{cite:ICML15RevGrad,cite:ICCV15SDT}. While performance was significantly improved, these state of the art methods may be restricted by the fact that the discriminative structures as well as complex multimode structures are not exploited for fine-grained alignment of different distributions.

Adversarial learning has been explored for generative modeling in Generative Adversarial Networks (GANs) \cite{cite:NIPS14AdversarialNet}. Recently, several difficulties of GANs have been addressed, e.g. ease training \cite{cite:Arxiv17WGAN,cite:ICLR17GAN}, avoid mode collapse \cite{cite:arXiv14CGAN,cite:ICLR17MGAN,cite:ICLR17UGAN}. In particular, Generative Multi-Adversarial Network (GMAN) \cite{cite:ICLR17GMAN} extends GANs to multiple discriminators including formidable adversary and forgiving teacher, which significantly eases model training and enhances distribution matching.

\begin{figure*}[tbp]
  \centering
  \includegraphics[width=0.8\textwidth]{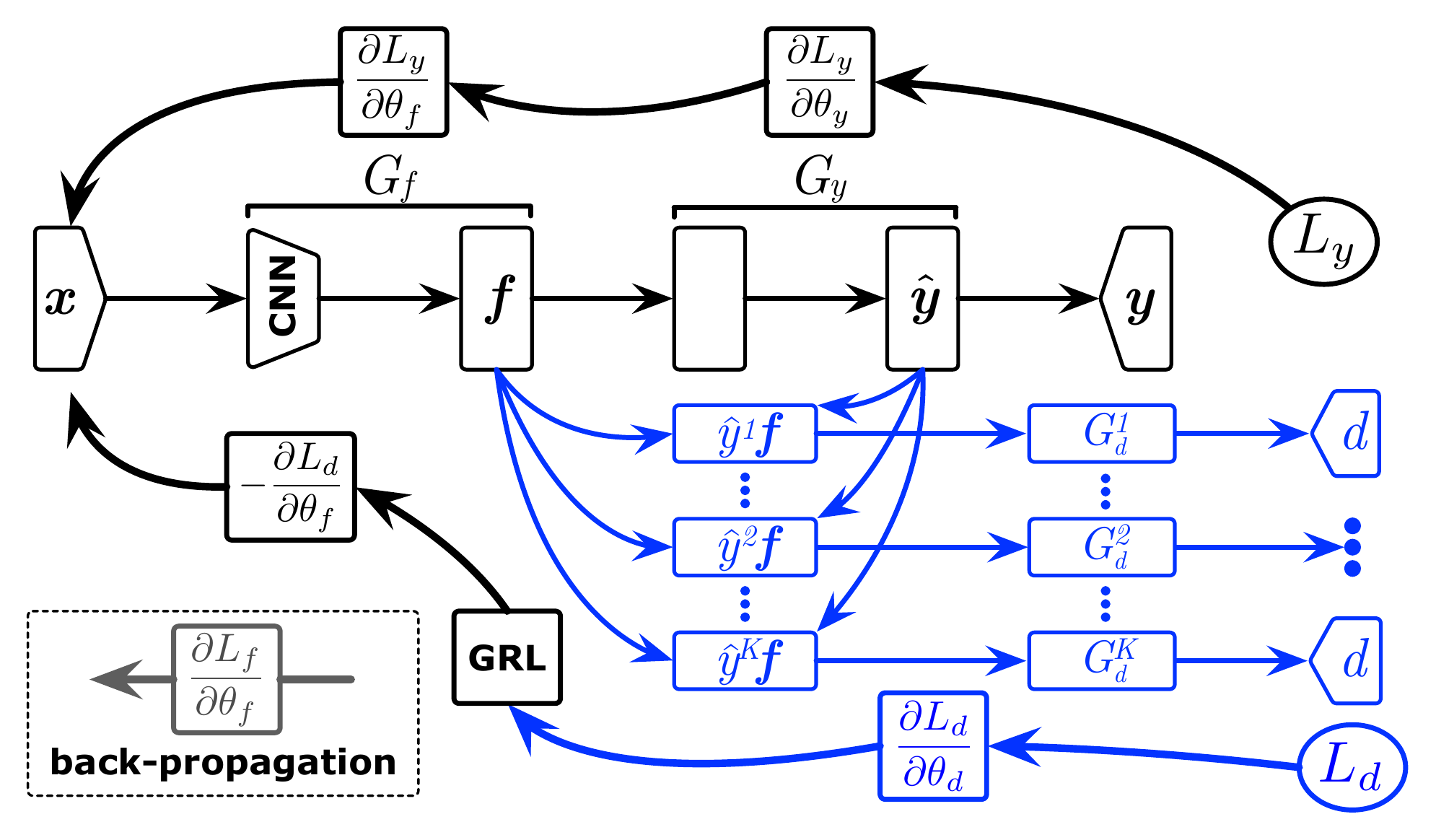}
  \caption{The architecture of the proposed Multi-Adversarial Domain Adaptation (MADA) approach, where $\mathbf{f}$ is the extracted deep features, ${\hat{\mathbf{y}}}$ is the predicted data label, and ${\hat{\mathbf{d}}}$ is the predicted domain label; $G_f$ is the feature extractor, $G_y$ and $L_y$ are the label predictor and its loss, $G_d^k$ and $L_d^k$ are the domain discriminator and its loss; GRL stands for Gradient Reversal Layer. The blue part shows the multiple adversarial networks (each for a class, $K$ in total) crafted in this paper. \emph{Best viewed in color.}}
  \label{fig:MADAfig}
\end{figure*}

\section{Multi-Adversarial Domain Adaptation}
In unsupervised domain adaptation, we are given a source domain $\mathcal{D}_s = \{(\mathbf{x}_i^s,{\bf y}^s_i)\}_{i=1}^{n_s}$ of $n_s$ labeled examples and a target domain ${{\cal D}_t} = \{ {\bf{x}}_j^t\} _{j = 1}^{{n_t}}$ of $n_t$ unlabeled examples. The source domain and target domain are sampled from joint distributions $P(\mathbf{X}^s, \mathbf{Y}^s)$ and $Q(\mathbf{X}^t, \mathbf{Y}^t)$ respectively, and note that $P \ne Q$. The goal of this paper is to design a deep neural network that enables learning of transfer features $\mathbf{f} = G_f\left( {\bf{x}} \right)$ and adaptive classifier $y = G_y\left( {\bf{f}} \right)$ to reduce the shifts in the joint distributions across domains, such that the target risk ${\Pr _{\left( {{\mathbf{x}},y} \right) \sim q}}\left[ {G_y \left( G_f({\mathbf{x}}) \right) \ne {\bf y}} \right]$ minimized by jointly minimizing source risk and distribution discrepancy by multi-adversarial domain adaptation.

There are two technical challenges to enabling domain adaptation: \textbf{(1)} enhancing \emph{positive} transfer by maximally matching the multimode structures underlying data distributions $P$ and $Q$ across domains, and \textbf{(2)} alleviating \emph{negative} transfer by preventing false alignment of different distribution modes across domains. These two challenges motivate the multi-adversarial domain adaptation approach.

\subsection{Domain Adversarial Network}
Domain adversarial networks have been successfully applied to transfer learning~\cite{cite:ICML15RevGrad,cite:ICCV15SDT} by extracting transferable features that can reduce the distribution shift between the source domain and the target domain. The adversarial learning procedure is a two-player game, where the first player is the domain discriminator $G_d$ trained to distinguish the source domain from the target domain, and the second player is the feature extractor $G_f$ fine-tuned simultaneously to confuse the domain discriminator.

To extract domain-invariant features $\mathbf{f}$, the parameters $\theta_f$ of feature extractor $G_f$ are learned by maximizing the loss of domain discriminator $G_d$, while the parameters $\theta_d$ of domain discriminator $G_d$ are learned by minimizing the loss of the domain discriminator. In addition, the loss of label predictor $G_y$ is also minimized. The objective of domain adversarial network \cite{cite:ICML15RevGrad} is the functional:
\begin{equation}\label{eqn:GRL}
\begin{aligned}
	C_{0} \left( {{\theta _f},{\theta _y},{\theta _d}} \right) &= \frac{1}{{{n_s}}}\sum\limits_{{{\mathbf{x}}_i} \in {\mathcal{D}_s}} {{L_y}\left( {{G_y}\left( {{G_f}\left( {{{\mathbf{x}}_i}} \right)} \right),{y_i}} \right)} \\
	&- \frac{\lambda }{n}\sum\limits_{{{\mathbf{x}}_i} \in \left( {{\mathcal{D}_s} \cup {\mathcal{D}_t}} \right)} {{L_d}\left( {{G_d}\left( {{G_f}\left( {{{\mathbf{x}}_i}} \right)} \right),{d_i}} \right)} ,
\end{aligned}
\end{equation}
where $n = n_s + n_t$ and $\lambda$ is a trade-off parameter between the two objectives that shape the features during learning.
After training convergence, the parameters $\hat\theta_f$, $\hat\theta_y$, $\hat\theta_d$ will deliver a saddle point of the functional~\eqref{eqn:GRL}: 
\begin{equation}\label{eqn:param1}
\begin{gathered}
     (\hat\theta_f, \hat\theta_y) = \arg \mathop {\min }\limits_{{\theta _f},{\theta _y}} C_0 \left( {{\theta _f},{\theta _y},{\theta _d}} \right), \\
     (\hat\theta_d) = \arg \mathop {\max }\limits_{{\theta_d}} C_0 \left( {{\theta _f},{\theta _y},{\theta _d}} \right).
\end{gathered}
\end{equation}
Domain adversarial networks \cite{cite:ICML15RevGrad,cite:ICCV15SDT} are the top-performing architectures for standard domain adaptation when the distributions of the source domain and target domain can be aligned successfully.

\subsection{Multi-Adversarial Domain Adaptation}
In practical domain adaptation problems, however, the data distributions of the source domain and target domain usually embody complex multimode structures, reflecting either the class boundaries in supervised learning or the cluster boundaries in unsupervised learning. Thus, previous domain adversarial adaptation methods that only match the data distributions without exploiting the multimode structures may be prone to either under transfer or negative transfer. Under transfer may happen when different modes of the distributions cannot be maximally matched. Negative transfer may happen when the corresponding modes of the distributions across domains are falsely aligned. To promote positive transfer and combat negative transfer, we should find a technology to reveal the multimode structures underlying distributions on which multi-adversarial domain adaptation can be performed.

To match the source and target domains upon the multimode structures underlying data distributions, we notice that the source domain labeled information provides strong signals to reveal the multimode structures. Therefore, we split the domain discriminator $G_d$ in Equation~\eqref{eqn:GRL} into $K$ class-wise domain discriminators $G_d^k, {k=1, \ldots, K}$, each is responsible for matching the source and target domain data associated with class $k$, as shown in Figure~\ref{fig:MADAfig}. Since target domain data are fully unlabeled, it is not easy to decide which domain discriminator $G_d^k$ is responsible for each target data point. Fortunately, 
we observe that the output of the label predictor ${\hat{\bf{y}}}_i = G_y(\mathbf{x}_i)$ to each data point $\mathbf{x}_i$ is a probability distribution over the label space of $K$ classes. This distribution well characterizes the probability of assigning $\mathbf{x}_i$ to each of the $K$ classes. Thus, it is a natural idea to use ${\hat{\bf{y}}}_i$ as the probability to indicate how much each data point $\mathbf{x}_i$ should be attended to the $K$ domain discriminators $G_d^k, {k=1, \ldots, K}$. The attention of each point $\mathbf{x}_i$ to a domain discriminator $G_d^k$ can be modeled by weighting its features $G_f(\mathbf{x}_i)$ with probability $\hat y_i^k$. Applying this to all $K$ domain discriminators $G_d^k, {k=1, \ldots, K}$ yields
\begin{equation}\label{eqn:Ld}
	{L_d} = \frac{1}{n}\sum\limits_{k = 1}^K {\sum\limits_{{{\mathbf{x}}_i} \in {\mathcal{D}_s} \cup {\mathcal{D}_t}} {L_d^k\left( {G_d^k\left( {\hat y_i^k{G_f}\left( {{{\mathbf{x}}_i}} \right)} \right),{d_i}} \right)} } ,
\end{equation}
where $G_d^k$ is the $k$-th domain discriminator while $L_d^k$ is its cross-entropy loss, and $d_i$ is the domain label of point $\mathbf{x}_i$. We note that the above strategy shares similar ideas with the attention mechanism.

Compared with the previous single-discriminator domain adversarial network in Equation~\eqref{eqn:GRL}, the proposed multi-adversarial domain adaptation network enables fine-grained adaptation where each data point $\mathbf{x}_i$ is matched only by those relevant domain discriminators according to its probability ${\hat{\bf{y}}}_i$. This fine-grained adaptation may introduce three benefits. \textbf{(1)} It avoids the hard assignment of each point to only one domain discriminator, which tends to be inaccurate for target domain data. \textbf{(2)} It circumvents negative transfer since each point is only aligned to the most relevant classes, while the irrelevant classes are filtered out by the probability and will not be included in the corresponding domain discriminators, hence avoiding false alignment of the discriminative structures in different distributions. \textbf{(3)} The multiple domain discriminators are trained with probability-weighted data points $\hat y_i^k G_f(\mathbf{x}_i)$, which naturally learn multiple domain discriminators with different parameters $\theta_d^k$; these domain discriminators with different parameters promote \emph{positive transfer} for each instance.

Integrating all things together, the objective of the Multi-Adversarial Domain Adaptation (MADA) is
\begin{equation}\label{eqn:MultiA}
\begin{aligned}
 C\left( {{\theta _f},{\theta _y},\theta _d^k|_{k = 1}^{K}} \right) & = \frac{1}{{{n_s}}}\sum\limits_{{{\mathbf{x}}_i} \in {\mathcal{D}_s}} {{L_y}\left( {{G_y}\left( {{G_f}\left( {{{\mathbf{x}}_i}} \right)} \right),{y_i}} \right)}  \\
  & - \frac{\lambda }{n}\sum\limits_{k = 1}^K {\sum\limits_{{{\mathbf{x}}_i} \in {\mathcal{D}}} {L_d^k\left( {G_d^k\left( {\hat y_i^k{G_f}\left( {{{\mathbf{x}}_i}} \right)} \right),{d_i}} \right)} },
\end{aligned}
\end{equation}  
where $n = n_s + n_t$, $\mathcal{D} = {\mathcal{D}_s} \cup {\mathcal{D}_t}$ and $\lambda$ is a hyper-parameter that trade-offs the two objectives in the unified optimization problem. The optimization problem is to find the parameters ${\hat\theta_f}$, ${\hat\theta_y}$ and ${\hat\theta_d^k}(k=1,2,...,K)$ that jointly satisfy
\begin{equation}\label{eqn:parameter1}
\begin{gathered}
     ({\hat\theta_f}, {\hat\theta_y}) = \arg \mathop {\min }\limits_{{\theta _f},{\theta _y}} C\left( {{\theta _f},{\theta _y},\theta _d^k|_{k = 1}^{K}} \right), \\
     ({\hat\theta_d^1},...,{\hat\theta_d^{K}}) = \arg \mathop {\max }\limits_{{\theta_d^1},...,{\theta_d^K}} C\left( {{\theta _f},{\theta _y},\theta _d^k|_{k = 1}^{K}} \right).
\end{gathered}
\end{equation}
The multi-adversarial domain adaptation (MADA) model simultaneously enhances \emph{positive} transfer by maximally matching the multimode structures underlying data distributions across domains, and circumvents \emph{negative} transfer by avoiding false alignment of the distribution modes across domains. 

\section{Experiments}
We evaluate the proposed multi-adversarial domain adaptation (MADA) model with state of the art transfer learning and deep learning methods. The codes, datasets and configurations will be available online at \url{github.com/thuml}.

\begin{table*}[!htbp]
  \centering
  \caption{Accuracy (\%) on \emph{Office-31} for unsupervised domain adaptation (AlexNet and ResNet)}
  \label{table:office31}
  \begin{small}
  \begin{tabular}{|c|c|c|c|c|c|c|c|}
    \hline
    Method & A $\rightarrow$ W & D $\rightarrow$ W & W $\rightarrow$ D & A $\rightarrow$ D & D $\rightarrow$ A & W $\rightarrow$ A & Avg \\
    \hline
    \hline
	AlexNet \cite{cite:NIPS12CNN} & 60.6$\pm$0.4 & 95.4$\pm$0.2 & 99.0$\pm$0.1 & 64.2$\pm$0.3 & 45.5$\pm$0.5 & 48.3$\pm$0.5 & 68.8\\
	TCA \cite{cite:TNN11TCA} & 59.0$\pm$0.0 & 90.2$\pm$0.0 & 88.2$\pm$0.0 & 57.8$\pm$0.0 & 51.6$\pm$0.0 & 47.9$\pm$0.0 & 65.8\\
	GFK \cite{cite:CVPR12GFK} & 58.4$\pm$0.0 & 93.6$\pm$0.0 & 91.0$\pm$0.0 & 58.6$\pm$0.0 & 52.4$\pm$0.0 & 46.1$\pm$0.0 & 66.7\\
	DDC \cite{cite:Arxiv14DDC} & 61.0$\pm$0.5 & 95.0$\pm$0.3 & 98.5$\pm$0.3 & 64.9$\pm$0.4 & 47.2$\pm$0.5 & 49.4$\pm$0.4 & 69.3\\
	DAN \cite{cite:ICML15DAN} & 68.5$\pm$0.3 & 96.0$\pm$0.1 & 99.0$\pm$0.1 & 66.8$\pm$0.2 & 50.0$\pm$0.4 & 49.8$\pm$0.3 & 71.7\\
    RTN \cite{cite:NIPS16RTN} & 73.3$\pm$0.2 & 96.8$\pm$0.2 & 99.6$\pm$0.1 & 71.0$\pm$0.2 & 50.5$\pm$0.3 & 51.0$\pm$0.1 & 73.7 \\
    RevGrad \cite{cite:ICML15RevGrad} & 73.0$\pm$0.5 & 96.4$\pm$0.3 & 99.2$\pm$0.3 & 72.3$\pm$0.3 & 52.4$\pm$0.4 & 50.4$\pm$0.5 & 74.1 \\
	\textbf{MADA} & \textbf{78.5}$\pm$0.2 & \textbf{99.8}$\pm$0.1 & \textbf{100.0}$\pm$.0 & \textbf{74.1}$\pm$0.1 & \textbf{56.0}$\pm$0.2 & \textbf{54.5}$\pm$0.3 & \textbf{77.1} \\
    \hline
    \hline
    ResNet \cite{cite:CVPR16DRL} & 68.4$\pm$0.2 & 96.7$\pm$0.1 & 99.3$\pm$0.1 & 68.9$\pm$0.2 & 62.5$\pm$0.3 & 60.7$\pm$0.3 & 76.1 \\
    TCA \cite{cite:TNN11TCA} & 74.7$\pm$0.0 & 96.7$\pm$0.0 & 99.6$\pm$0.0 & 76.1$\pm$0.0 & 63.7$\pm$0.0 & 62.9$\pm$0.0 & 79.3 \\
    GFK \cite{cite:CVPR12GFK} & 74.8$\pm$0.0 & 95.0$\pm$0.0 & 98.2$\pm$0.0 & 76.5$\pm$0.0 & 65.4$\pm$0.0 & 63.0$\pm$0.0 & 78.8 \\
    DDC \cite{cite:Arxiv14DDC} & 75.8$\pm$0.2 & 95.0$\pm$0.2 & 98.2$\pm$0.1 & 77.5$\pm$0.3 & 67.4$\pm$0.4 & 64.0$\pm$0.5 & 79.7 \\
    DAN \cite{cite:ICML15DAN} & 83.8$\pm$0.4 & 96.8$\pm$0.2 & 99.5$\pm$0.1 & 78.4$\pm$0.2 & 66.7$\pm$0.3 & 62.7$\pm$0.2 & 81.3 \\
    RTN \cite{cite:NIPS16RTN} & 84.5$\pm$0.2 & 96.8$\pm$0.1 & 99.4$\pm$0.1 & 77.5$\pm$0.3 & 66.2$\pm$0.2 & 64.8$\pm$0.3 & 81.6 \\
    RevGrad \cite{cite:ICML15RevGrad} & 82.0$\pm$0.4 & 96.9$\pm$0.2 & 99.1$\pm$0.1 & 79.7$\pm$0.4 & 68.2$\pm$0.4 & \textbf{67.4}$\pm$0.5 & 82.2 \\
    \textbf{MADA} & \textbf{90.0}$\pm$0.1 & \textbf{97.4}$\pm$0.1 & \textbf{99.6}$\pm$0.1 & \textbf{87.8}$\pm$0.2 & \textbf{70.3}$\pm$0.3 & {66.4}$\pm$0.3 & \textbf{85.2} \\
    \hline
  \end{tabular}
  \end{small}
\end{table*}

\subsection{Setup}
\textbf{Office-31} \cite{cite:ECCV10Office} is a standard benchmark for visual domain adaptation, comprising 4,652 images and 31 categories collected from three distinct domains: \textit{Amazon} (\textbf{A}), which contains images downloaded from \url{amazon.com}, \textit{Webcam} (\textbf{W}) and \textit{DSLR} (\textbf{D}), which contain images respectively taken by web camera and digital SLR camera with different environments. We evaluate all methods across three transfer tasks \textbf{A} $\rightarrow$ \textbf{W}, \textbf{D} $\rightarrow$ \textbf{W} and \textbf{W} $\rightarrow$ \textbf{D}, which are widely used by previous deep transfer learning methods \cite{cite:Arxiv14DDC,cite:ICML15RevGrad}, and another three transfer tasks \textbf{A} $\rightarrow$ \textbf{D}, \textbf{D} $\rightarrow$ \textbf{A} and \textbf{W} $\rightarrow$ \textbf{A} as used in \cite{cite:ICML15DAN,cite:ICCV15SDT,cite:NIPS16RTN}.

\textbf{ImageCLEF-DA}\footnote{\url{http://imageclef.org/2014/adaptation}} is a benchmark dataset for ImageCLEF 2014 domain adaptation challenge, which is organized by selecting the 12 common categories shared by the following three public datasets, each is considered as a domain: \textit{Caltech-256} (\textbf{C}), \textit{ImageNet ILSVRC 2012} (\textbf{I}), and \textit{Pascal VOC 2012} (\textbf{P}). 
The 12 common categories are {aeroplane}, {bike}, {bird}, {boat}, {bottle}, {bus}, {car}, {dog}, {horse}, {monitor}, {motorbike}, and {people}. There are 50 images in each category and 600 images in each domain.
We use all domain combinations and build 6 transfer tasks: \textbf{I} $\rightarrow$ \textbf{P}, \textbf{P} $\rightarrow$ \textbf{I}, \textbf{I} $\rightarrow$ \textbf{C}, \textbf{C} $\rightarrow$ \textbf{I}, \textbf{C} $\rightarrow$ \textbf{P}, and \textbf{P} $\rightarrow$ \textbf{C}. Different from the \emph{Office-31} dataset where different domains are of different sizes, the three domains in this dataset are of equal size, making it a good alternative dataset.

We compare the proposed multi-adversarial domain adaptation (\textbf{MADA)} with both shallow and deep transfer learning methods: Transfer Component Analysis (\textbf{TCA}) \cite{cite:TNN11TCA}, Geodesic Flow Kernel (\textbf{GFK}) \cite{cite:CVPR12GFK}, Deep Domain Confusion (\textbf{DDC}) \cite{cite:Arxiv14DDC}, Deep Adaptation Network (\textbf{DAN}) \cite{cite:ICML15DAN}, Residual Transfer Network (\textbf{RTN}) \cite{cite:NIPS16RTN}, and Reverse Gradient (\textbf{RevGrad}) \cite{cite:ICML15RevGrad}. TCA learns a shared feature space by Kernel PCA with linear-MMD penalty. GFK interpolates across an infinite number of intermediate subspaces to bridge the source and target subspaces. For these shallow transfer methods, we adopt SVM as the base classifier. DDC maximizes domain confusion by adding to deep networks a single adaptation layer that is regularized by linear-kernel MMD. DAN learns transferable features by embedding deep features of multiple domain-specific layers to reproducing kernel Hilbert spaces (RKHSs) and matching different distributions optimally using multi-kernel MMD. RTN jointly learns transferable features and adapts different source and target classifiers via deep residual learning \cite{cite:CVPR16DRL}. RevGrad enables domain adversarial learning \cite{cite:NIPS14AdversarialNet} by adapting a single layer of deep networks, which matches the source and target domains by making them indistinguishable for a domain discriminator.

We follow standard evaluation protocols for unsupervised domain adaptation \cite{cite:ICML15DAN,cite:ICML15RevGrad}. For both \emph{Office-31} and \emph{ImageCLEF-DA} datasets, we use all labeled source examples and all unlabeled target examples. We compare the average classification accuracy of each method on three random experiments, and report the standard error of the classification accuracies by different experiments of the same transfer task. For all baseline methods, we either follow their original model selection procedures, or conduct \emph{transfer cross-validation} \cite{cite:ECML10TCV} if their model selection strategies are not specified. We also adopt transfer cross-validation \cite{cite:ECML10TCV} to select parameter $\lambda$ for the MADA models. Fortunately, our models perform very stably under different parameter values, thus we fix $\lambda = 1$ throughout all experiments. 
For MMD-based methods (TCA, DDC, DAN, and RTN), we use Gaussian kernel with bandwidth set to the median pairwise squared distances on the training data, i.e. median trick \cite{cite:JMLR12MMD,cite:ICML15DAN}. 
We examine the influence of deep representations for domain adaptation by exploring \textbf{AlexNet} \cite{cite:NIPS12CNN} and \textbf{ResNet} \cite{cite:CVPR16DRL} as base architectures for learning deep representations. For shallow methods, we follow DeCAF \cite{cite:ICML14DeCAF} and use as deep representations the activations of the $fc7$ (AlexNet) and $pool5$ (ResNet) layers. 

We implement all deep methods based on the \textbf{Caffe} \cite{cite:MM14Caffe} framework, and fine-tune from AlexNet \cite{cite:NIPS12CNN} and ResNet \cite{cite:CVPR16DRL} models pre-trained on the ImageNet dataset \cite{cite:Arxiv14ImageNet}. We fine-tune all convolutional and pooling layers and train the classifier layer via back propagation. Since the classifier is trained from scratch, we set its learning rate to be 10 times that of the lower layers. 
We employ the mini-batch stochastic gradient descent (SGD) with momentum of 0.9 and the learning rate strategy implemented in RevGrad \cite{cite:ICML15RevGrad}: the learning rate is not selected by a grid search due to high computational cost---it is adjusted during SGD using these formulas: ${\eta _p} = \frac{{{\eta _0}}}{{{{\left( {1 + \alpha p} \right)}^\beta }}}$, where $p$ is the training progress linearly changing from $0$ to $1$, $\eta_0 = 0.01, \alpha=10$ and $\beta=0.75$, which is optimized to promote convergence and low error on source domain. To suppress noisy activations at the early stages of training, instead of fixing parameter $\lambda$, we gradually change it by multiplying $\frac{2}{{1 + \exp \left( { - \delta p} \right)}} - 1$, where $\delta = 10$ \cite{cite:ICML15RevGrad}. This progressive training strategy significantly stabilizes parameter sensitivity of the proposed approach.

\begin{table*}[!htbp]
  \centering
  \caption{Accuracy (\%) on \emph{ImageCLEF-DA} for unsupervised domain adaptation (AlexNet and ResNet)}
  \label{table:imageclefda}
  \begin{small}
  \begin{tabular}{|c|c|c|c|c|c|c|c|}
    \hline
    Method & I $\rightarrow$ P & P $\rightarrow$ I & I $\rightarrow$ C & C $\rightarrow$ I & C $\rightarrow$ P & P $\rightarrow$ C & Avg \\
    \hline
    \hline
    AlexNet \cite{cite:NIPS12CNN} & 66.2$\pm$0.2 & 70.0$\pm$0.2 & 84.3$\pm$0.2 & 71.3$\pm$0.4 & 59.3$\pm$0.5 & 84.5$\pm$0.3 & 73.9 \\
    DAN \cite{cite:ICML15DAN} & 67.3$\pm$0.2 & 80.5$\pm$0.3 & 87.7$\pm$0.3 & 76.0$\pm$0.3 & 61.6$\pm$0.3 & 88.4$\pm$0.2 & 76.9 \\
    RTN \cite{cite:NIPS16RTN} & 67.4$\pm$0.3 & 82.3$\pm$0.3 & 89.5$\pm$0.4 & 78.0$\pm$0.2 & 63.0$\pm$0.2 & 90.1$\pm$0.1 & 78.4 \\
    RevGrad \cite{cite:ICML15RevGrad} & 66.5$\pm$0.5 & 81.8$\pm$0.4 & 89.0$\pm$0.5 & 79.8$\pm$0.5 & 63.5$\pm$0.4 & 88.7$\pm$0.4 & 78.2 \\
    \textbf{MADA} & \textbf{68.3}$\pm$0.3 & \textbf{83.0}$\pm$0.1 & \textbf{91.0}$\pm$0.2 & \textbf{80.7}$\pm$0.2 & \textbf{63.8}$\pm$0.2 & \textbf{92.2}$\pm$0.3 & \textbf{79.8} \\
    \hline
    \hline
    ResNet \cite{cite:CVPR16DRL} & 74.8$\pm$0.3 & 83.9$\pm$0.1 & 91.5$\pm$0.3 & 78.0$\pm$0.2 & 65.5$\pm$0.3 & 91.2$\pm$0.3 & 80.7 \\
    DAN \cite{cite:ICML15DAN} & 75.0$\pm$0.4 & 86.2$\pm$0.2 & 93.3$\pm$0.2 & 84.1$\pm$0.4 & 69.8$\pm$0.4 & 91.3$\pm$0.4 & 83.3 \\
    RTN \cite{cite:NIPS16RTN} & \textbf{75.6}$\pm$0.3 & 86.8$\pm$0.1 & 95.3$\pm$0.1 & 86.9$\pm$0.3 & 72.7$\pm$0.3 & 92.2$\pm$0.4 & 84.9 \\
    RevGrad \cite{cite:ICML15RevGrad} & 75.0$\pm$0.6 & 86.0$\pm$0.3 & \textbf{96.2}$\pm$0.4 & 87.0$\pm$0.5 & 74.3$\pm$0.5 & 91.5$\pm$0.6 & 85.0 \\
    \textbf{MADA} & {75.0}$\pm$0.3 & \textbf{87.9}$\pm$0.2 & 96.0$\pm$0.3 & \textbf{88.8}$\pm$0.3 & \textbf{75.2}$\pm$0.2 & \textbf{92.2}$\pm$0.3 & \textbf{85.8} \\
    \hline
  \end{tabular}
  \end{small}
\end{table*}

\begin{table*}[!htbp]
  \centering
  \caption{Accuracy (\%) on \emph{Office-31} for domain adaptation from 31 classes to 25 classes (AlexNet)}
  \label{table:office25}
  \begin{small}
  \begin{tabular}{|c|c|c|c|c|c|c|c|}
    \hline
    Method & A $\rightarrow$ W & D $\rightarrow$ W & W $\rightarrow$ D & A $\rightarrow$ D & D $\rightarrow$ A & W $\rightarrow$ A & Avg \\
    \hline
    \hline
	AlexNet \cite{cite:NIPS12CNN} & 58.2$\pm$0.4 & 95.9$\pm$0.2 & 99.0$\pm$0.1 & 60.4$\pm$0.3 & 49.8$\pm$0.5 & 47.3$\pm$0.5 & 68.4\\
    RevGrad \cite{cite:ICML15RevGrad} & 65.1$\pm$0.5 & 91.7$\pm$0.3 & 97.1$\pm$0.3 & 60.6$\pm$0.3 & 42.1$\pm$0.4 & 42.9$\pm$0.5 & 66.6 \\
	\textbf{MADA} & \textbf{70.8}$\pm$0.2 & \textbf{96.6}$\pm$0.1 & \textbf{99.5}$\pm$.0 & \textbf{69.6}$\pm$0.1 & \textbf{51.4}$\pm$0.2 & \textbf{54.2}$\pm$0.3 & \textbf{73.7} \\
    \hline
  \end{tabular}
  \end{small}
\end{table*}

\subsection{Results}
The classification accuracy results on the \textit{Office-31} dataset for unsupervised domain adaptation based on AlexNet and ResNet are shown in Table \ref{table:office31}. For fair comparison, the results of DAN \cite{cite:ICML15DAN}, RTN \cite{cite:NIPS16RTN}, and RevGrad \cite{cite:ICML15RevGrad} are directly reported from their original papers. MADA outperforms all comparison methods on most transfer tasks. It is noteworthy that MADA promotes the classification accuracies substantially on hard transfer tasks, e.g. \textbf{A $\rightarrow$ W}, \textbf{A $\rightarrow$ D}, \textbf{D $\rightarrow$ A}, and \textbf{W $\rightarrow$ A}, where the source and target domains are substantially different, and produce comparable classification accuracies on easy transfer tasks, \textbf{D $\rightarrow$ W} and \textbf{W $\rightarrow$ D}, where the source and target domains are similar \cite{cite:ECCV10Office}. 
The three domains in the \textit{ImageCLEF-DA} dataset are balanced in each category. As reported in Table \ref{table:imageclefda}, the MADA approach outperforms the comparison methods on most transfer tasks.
The encouraging results highlight the importance of multi-adversarial domain adaptation in deep neural networks, and suggest that MADA is able to learn more transferable representations for effective domain adaptation.

The experimental results reveal several insightful observations. \textbf{(1)} Standard deep learning methods (AlexNet and ResNet) either outperform or underperform traditional shallow transfer learning methods (TCA and GFK) using deep features as input. This confirms the current practice that deep networks, even the extremely deep ones (ResNet), can learn abstract feature representations that only reduce but not remove the cross-domain discrepancy \cite{cite:NIPS14CNN}. \textbf{(2)} Deep transfer learning methods substantially outperform both standard deep learning methods and traditional shallow transfer learning methods with deep features as input. This validates that explicitly reducing the cross-domain discrepancy by embedding domain-adaptation modules into deep networks (DDC, DAN, RTN, and RevGrad) can learn more transferable features. \textbf{(3)} MADA substantially outperforms previous methods based on either multilayer adaptation (DAN), semi-supervised adaptation (RTN), and domain adversarial training (RevGrad). Although both MADA and RevGrad \cite{cite:ICML15RevGrad} perform domain adversarial adaptation, the improvement from RevGrad to MADA is crucial for domain adaptation: RevGrad matches data distributions across domains without exploiting the complex multimode structures; MADA enables domain adaptation by making the source and target domains indistinguishable multiple domain discriminators, each responsible for matching the source and target data associated with the same class, which can essentially reduce the shift in the data distributions of complex multimode structures.

\begin{figure*}[h]
  \centering
  \subfigure[RevGrad: source=\textbf{A}]{
    \includegraphics[width=0.22\textwidth]{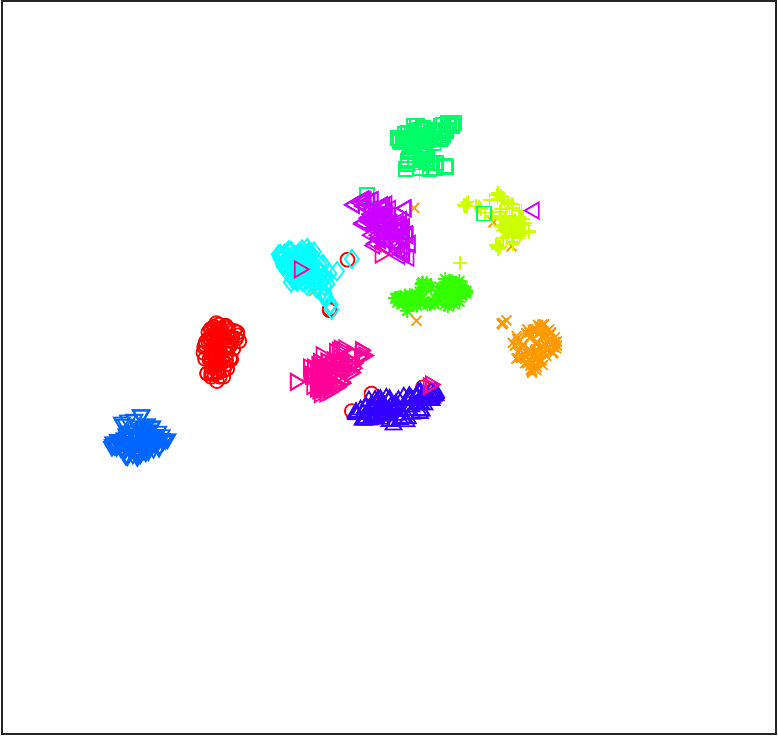}
    \label{fig:RevGrad1}
  }\hfil
  \subfigure[RevGrad: target=\textbf{W}]{
    \includegraphics[width=0.22\textwidth]{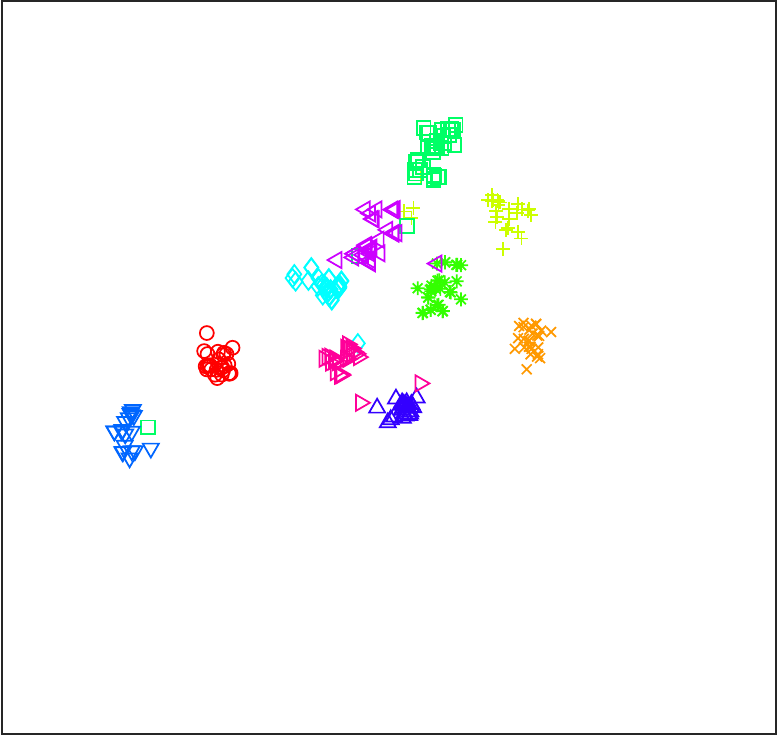}
    \label{fig:RevGrad2}
  }\hfil
  \subfigure[MADA: source=\textbf{A}]{
    \includegraphics[width=0.22\textwidth]{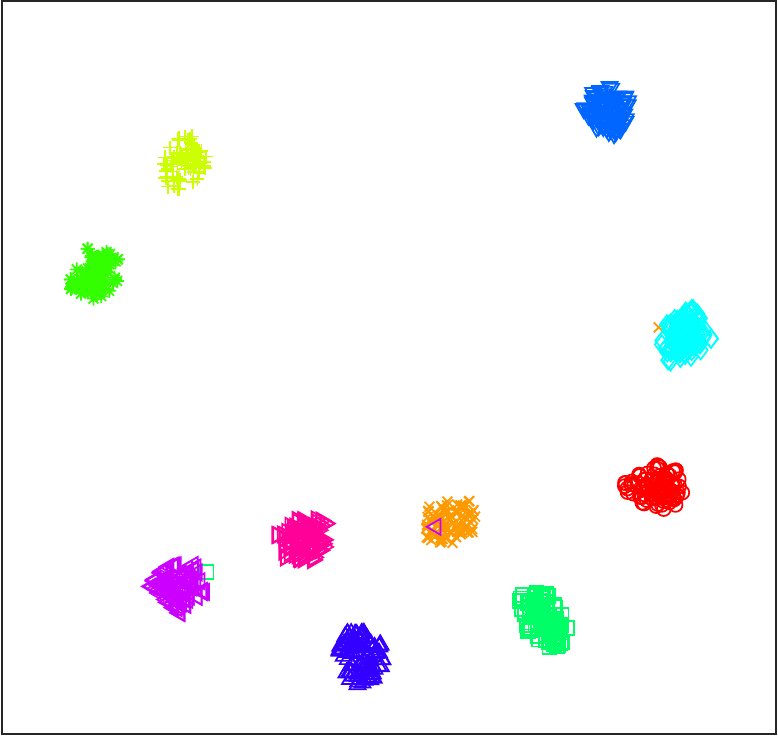}
    \label{fig:MADA1}
  }\hfil
  \subfigure[MADA: target=\textbf{W}]{
    \includegraphics[width=0.22\textwidth]{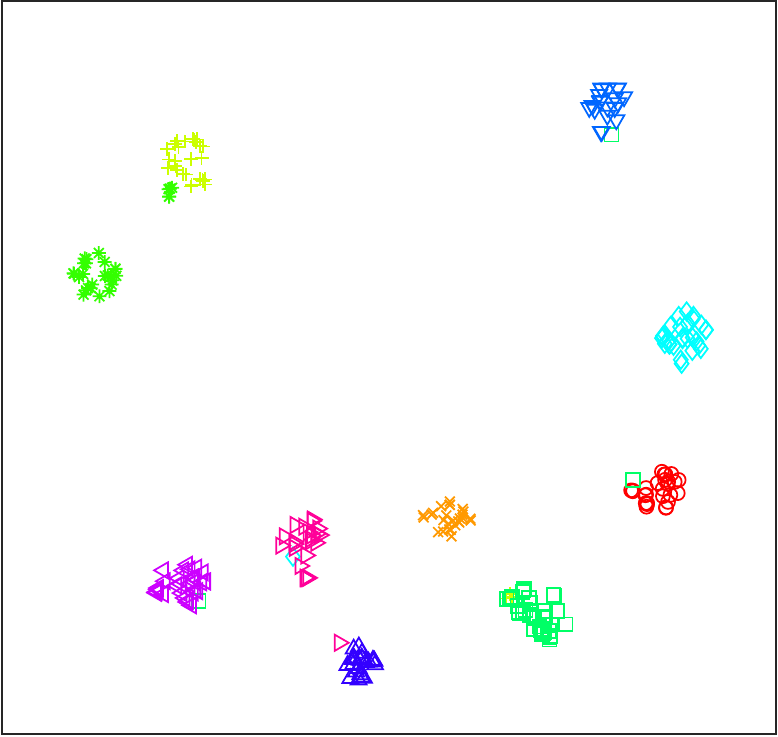}
    \label{fig:MADA2}
  }
  \caption{The t-SNE visualization of deep features extracted by RevGrad (a)(b) and MADA (c)(d).}
\end{figure*}

\begin{figure*}[h]
  \centering
  \subfigure[Sharing Strategies]{
    \includegraphics[width=0.3\textwidth]{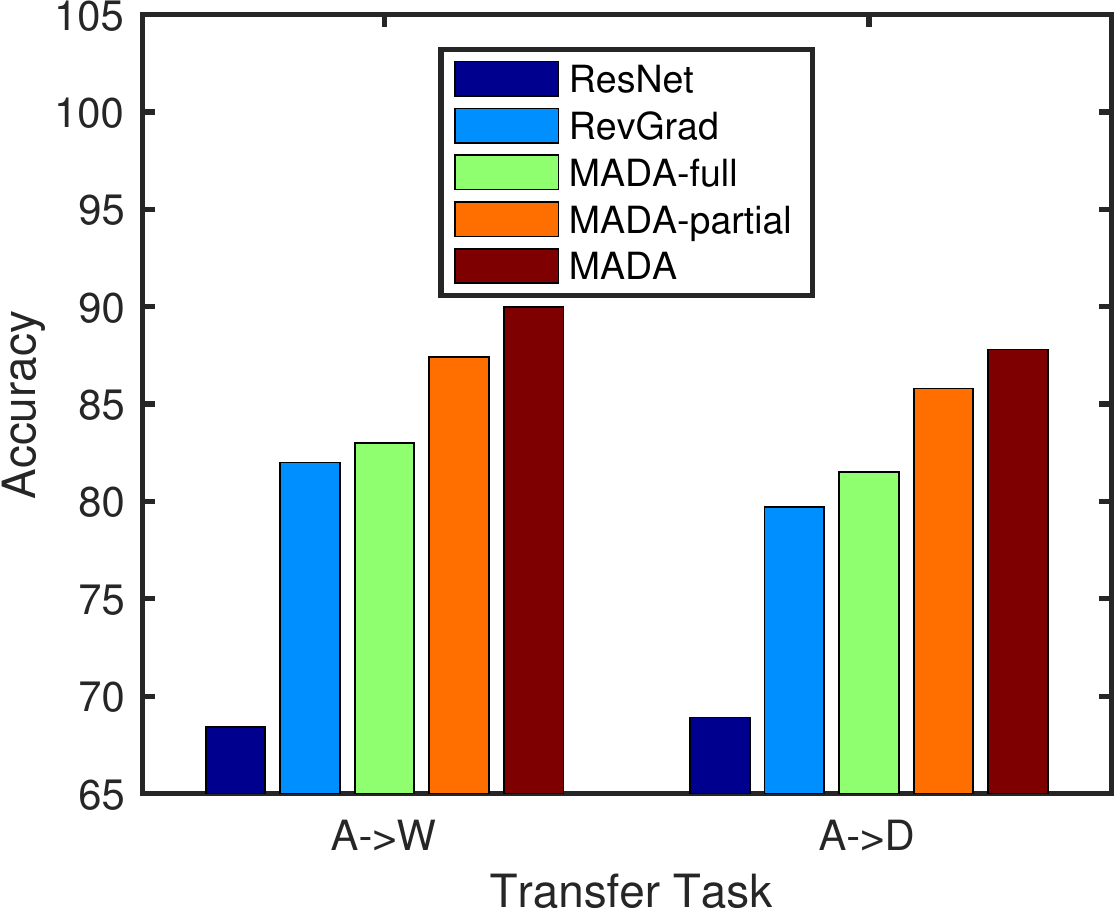}
    \label{fig:sharing}
  }\hfil
  \subfigure[${\cal A}$-distance]{
    \includegraphics[width=0.3\textwidth]{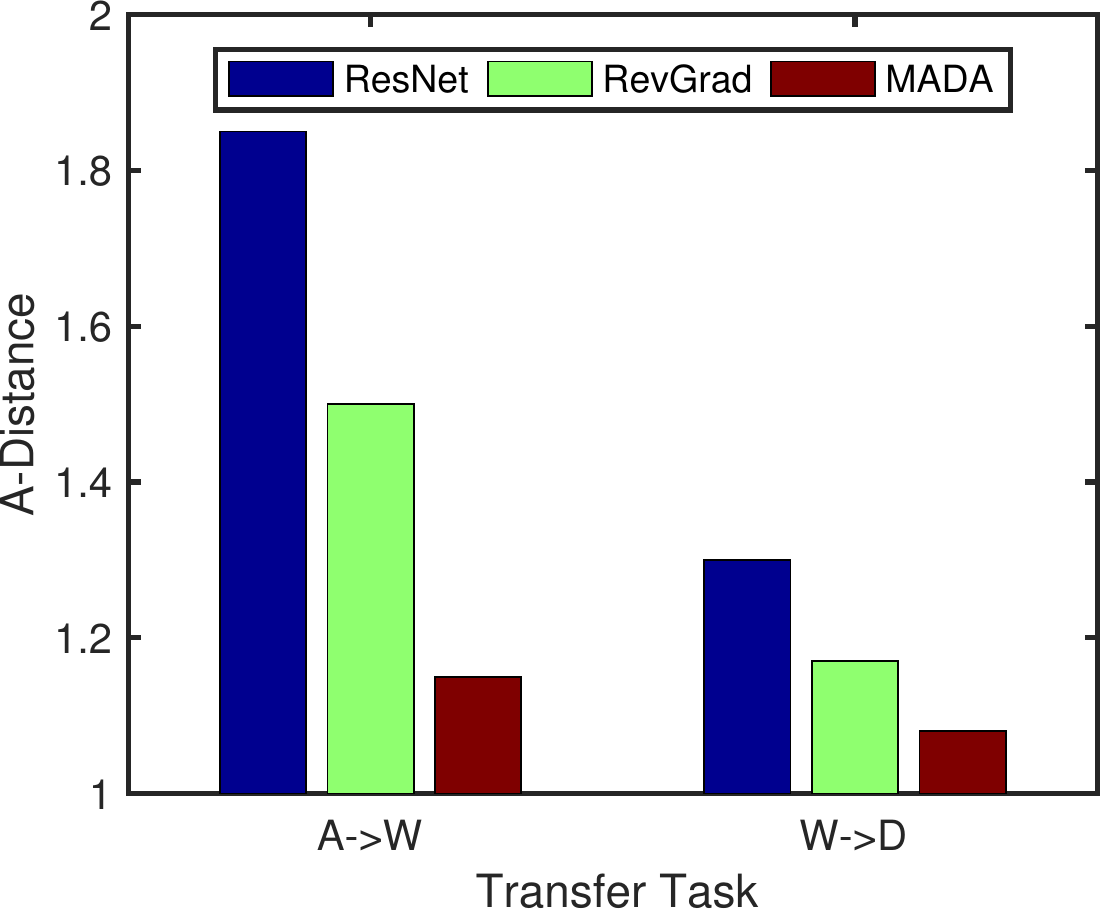}
    \label{fig:Adist}
  }\hfil
  \subfigure[Convergence]{
    \includegraphics[width=0.3\textwidth]{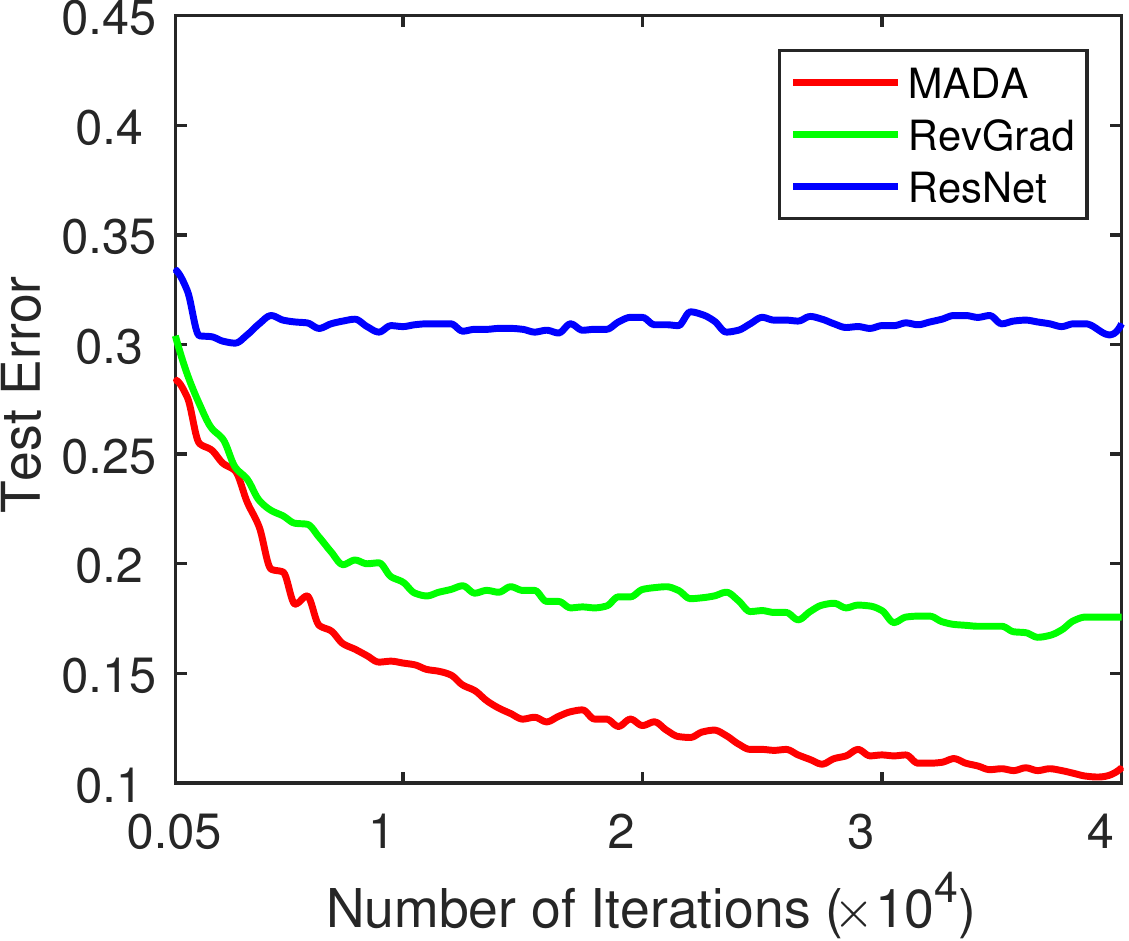}
    \label{fig:error}
  }\hfil
  \caption{Empirical analysis: (a) Sharing strategies, (b) ${\cal A}$-distance, and (c) Convergence performance.}
\end{figure*}

Negative transfer is an important technical bottleneck for successful domain adaptation. Negative transfer is more likely to happen when the source domain is substantially larger than the target domain, in which there exist many  source data points that are irrelevant to the target domain. To evaluate the robustness against negative transfer, we randomly remove 6 classes from all transfer tasks constructed from the \emph{Office-31} dataset. For example, we perform domain adaptation on transfer task \textbf{A 31 $\rightarrow$ W 25}, where the source domain \textbf{A} has 31 classes but the target domain \textbf{W} has only 25 classes. In this more general and challenging scenario, we observe from Table~\ref{table:office25} that the top-performing adversarial adaptation method, RevGrad, significantly underperforms standard AlexNet on most transfer tasks. This is an evidence of the negative transfer difficulty. The proposed MADA approach significantly exceeds the performance of both AlexNet and RevGrad, and successfully avoids the negative transfer trap. These positive results imply that the multi-adversarial adaptation can alleviate negative transfer.

\subsection{Analysis}
\textbf{Feature Visualization:}
We go deeper into the feature transferability by visualizing in Figures~\ref{fig:RevGrad1}--\ref{fig:MADA2} the network activations of task \textbf{A} $\rightarrow$ \textbf{W} (10 classes) learned by RevGrad (the bottleneck layer $fcb$) and MADA (the bottleneck layer $fcb$) respectively using t-SNE embeddings \cite{cite:ICML14DeCAF}. The visualization results reveal several interesting observations. \textbf{(1)} Under RevGrad features, the source and target domains are made indistinguishable; however, different categories are not well discriminated clearly. The reason is that domain adversarial learning is performed only at the feature layer $fcb$, while the discriminative information is not taken into account by the domain adversary. \textbf{(2)} Under MADA features, not only the source and target domains are made more indistinguishable but also different categories are made more discriminated, which leads to the best adaptation accuracy. This superior results benefit from the integration of discriminative information into multiple domain discriminators, which enables matching of complex multimode structures of the source and target data distributions.

\textbf{Sharing Strategies:}
Besides the proposed multi-adversarial strategy, one may consider other sharing strategies for multiple domain discriminators. For example, one can consider sharing all network parameters in the multiple domain discriminators, which is similar to previous domain adversarial adaptation methods with single domain discriminator; or consider sharing only a fraction of the network parameters for more flexibility. To examine different sharing strategies, we compare different variants of MADA: \textbf{MADA-full}, which shares all parameters of the multiple domain discriminator networks; \textbf{MADA-partial}, which shares only the lowest layers of the multiple discriminator networks. The accuracy results of tasks \textbf{A $\rightarrow$ W} and \textbf{A $\rightarrow$ D} in Figure~\ref{fig:sharing} reveal that the transfer performance decreases when we share more parameters of multiple discriminators. This confirms our motivation that multiple domain discriminators are necessary to establish fine-grained distribution alignment.

\textbf{Distribution Discrepancy:}
The domain adaptation theory~\cite{cite:ML10DAT,cite:COLT09DAT} suggests $\mathcal{A}$-distance as a measure of cross-domain discrepancy, which, together with the source risk, will bound the target risk. The proxy ${\cal{A}}$-distance is defined as ${d_{\cal A}} = 2\left( {1 - 2\epsilon } \right)$, where $\epsilon$ is the generalization error of a classifier (e.g. kernel SVM) trained on the binary task of discriminating source and target. Figure~\ref{fig:Adist} shows ${d_{\cal A}}$ on tasks \textbf{A} $\rightarrow$ \textbf{W}, \textbf{W} $\rightarrow$ \textbf{D} with features of ResNet, RevGrad, and MADA. We observe that ${d_{\cal A}}$ using MADA features is much smaller than ${d_{\cal A}}$ using ResNet and RevGrad features, which suggests that MADA features can reduce the cross-domain gap more effectively. As domains \textbf{W} and \textbf{D} are similar, ${d_{\cal A}}$ of task \textbf{W} $\rightarrow$ \textbf{D} is smaller than that of \textbf{A} $\rightarrow$ \textbf{W}, which well explains better accuracy of \textbf{W} $\rightarrow$ \textbf{D}.

\textbf{Convergence Performance:}
Since MADA involves alternating optimization procedures, we testify the convergence performance with ResNet and RevGrad. Figure~\ref{fig:error} demonstrates the test errors of different methods on task \textbf{A $\rightarrow$ W}, which suggests that MADA has similarly stable convergence performance as RevGrad while significantly outperforming RevGrad in the whole process of convergence. Also, the computational complexity of MADA is similar to RevGrad since the multiple domain discriminators only occupy a small fraction of the overall computational complexity.

\section{Conclusion}
This paper presented a novel multi-adversarial domain adaptation approach to enable effective deep transfer learning. Unlike previous domain adversarial adaptation methods that only match the feature distributions across domains without exploiting the complex multimode structures, the proposed approach further exploits the discriminative structures to enable fine-grained distribution alignment in a multi-adversarial adaptation framework, which can simultaneously promote positive transfer and circumvent negative transfer. Experiments show state of the art results of the proposed approach.

\section*{Acknowledgments}
This work was supported by the National Key Research and Development Program of China (2016YFB1000701), National Natural Science Foundation of China (61772299, 61325008, 61502265, 61672313) and Tsinghua National Laboratory (TNList) Key Project.

\begin{small}
	\bibliography{MADA}
	\bibliographystyle{aaai}
\end{small}

\end{document}